\documentclass[a4paper]{article}

\usepackage{INTERSPEECH2022}

\newcommand\bertbase{BERT$_{\small \textsc{BASE}}$}

\newcommand\robertabase{RoBERTa$_{\small \textsc{BASE}}$}
\newcommand\xlnetlarge{XLNet$_{\small \textsc{LARGE}}$}
\newcommand\tbase{T5$_{\small \textsc{BASE}}$}
\newcommand\tlarge{T5$_{\small \textsc{LARGE}}$}
\newcommand\txxl{T5$_{\small \textsc{XXL}}$}

\title{A Multi-Task BERT Model for Schema-Guided Dialogue State Tracking}
\name{Eleftherios Kapelonis$^1$, Efthymios Georgiou$^{1,2}$, Alexandros Potamianos$^1$}
\address{
  $^1$School of ECE, National Technical University of Athens, Athens, Greece\\
  $^2$Institute for Language and Speech Processing, Athena Research Center, Athens, Greece}
\email{lkapelonis@gmail.com, efthygeo@mail.ntua.gr, potam@central.ntua.gr}

\begin{document}

\maketitle

\begin{abstract}
Task-oriented dialogue systems often employ a Dialogue State Tracker (DST) to successfully complete conversations.
Recent state-of-the-art DST implementations rely on schemata of diverse services to improve model robustness and handle zero-shot generalization to new domains \cite{rastogi2020scalable}, however such methods \cite{ma2020endtoend, zhao2022descriptiondriven} typically require multiple
large scale transformer models and long input sequences to perform well.
We propose a single multi-task BERT-based model that jointly solves the three DST tasks of intent prediction, requested slot prediction and slot filling.
Moreover, we propose an efficient and parsimonious encoding of the dialogue history and service schemata that is shown to further improve performance.
Evaluation on the SGD dataset shows that our approach outperforms the baseline SGP-DST by a large margin and performs well compared to the state-of-the-art, while being significantly more computationally efficient.
Extensive ablation studies are performed to examine the contributing factors to the success of our model.
\end{abstract}

\section{Introduction}

Task-oriented dialogue is an important and active research area that has attracted a lot of attention in both academia and industry.
The aim of task-oriented dialogue systems is to assist users in accomplishing daily activities like reserving a restaurant, booking tickets etc.
An important component of a task-oriented dialogue system is the Dialogue State Tracker (DST) which tracks the user goal over multiple turns of dialogue.
Based on a spoken utterance and the dialogue history, the DST predicts the dialogue state which represents the user goal.
The predicted dialogue state is then used by other components to retrieve elements from a database, perform the actions requested by the user and respond accordingly \cite{DBLP:journals/corr/abs-2003-07490}.

Both single-domain \cite{williams-etal-2013-dialog, henderson-etal-2014-second, DBLP:journals/corr/BordesW16, DBLP:journals/corr/WenGMRSUVY16} and multi-domain \cite{budzianowski2018multiwoz, eric2019multiwoz, zang2020multiwoz, han2021multiwoz, ye2021multiwoz} approaches have been used for DST training.
The main DST's tasks are to predict the active user intent (intent prediction), the slots that are requested by the user (requested slot prediction) and the values for slots given by the user until the turn (slot filling) \cite{DBLP:journals/corr/abs-2002-01359}.
Early neural methods use slot-dependent architectures \cite{Henderson2014WordBasedDS, DBLP:journals/corr/MrksicSWTY16}, training different parameters for every slot.
In order to improve scalability and performance, slot-independent methods were proposed \cite{DBLP:journals/corr/abs-1712-10224, DBLP:journals/corr/abs-1810-09587} which share parameters between all slots.

Motivated by the ever-increasing number of diverse services used by commercial task-oriented systems, the schema-guided paradigm was developed \cite{rastogi2020scalable}. Services or dialogue domains are defined by their corresponding \textit{schema}, a structured ontology, which is usually a set of the supported intents and slots.
Schema-guided approaches \cite{rastogi2020scalable, eric2019multiwoz, DBLP:journals/corr/abs-2010-11853} often include a natural language description of the schema elements, e.g., in the SGD dataset the schema for the service Restaurants\_1 has a slot with name ``party\_size'' and description ``Party size for a reservation''. An important goal of schema-based approaches is scalability and generalization, i.e., to build systems that are capable of handling completely new domains and services.

\begin{figure}[t]
  \centering
  \includegraphics[width=\linewidth]{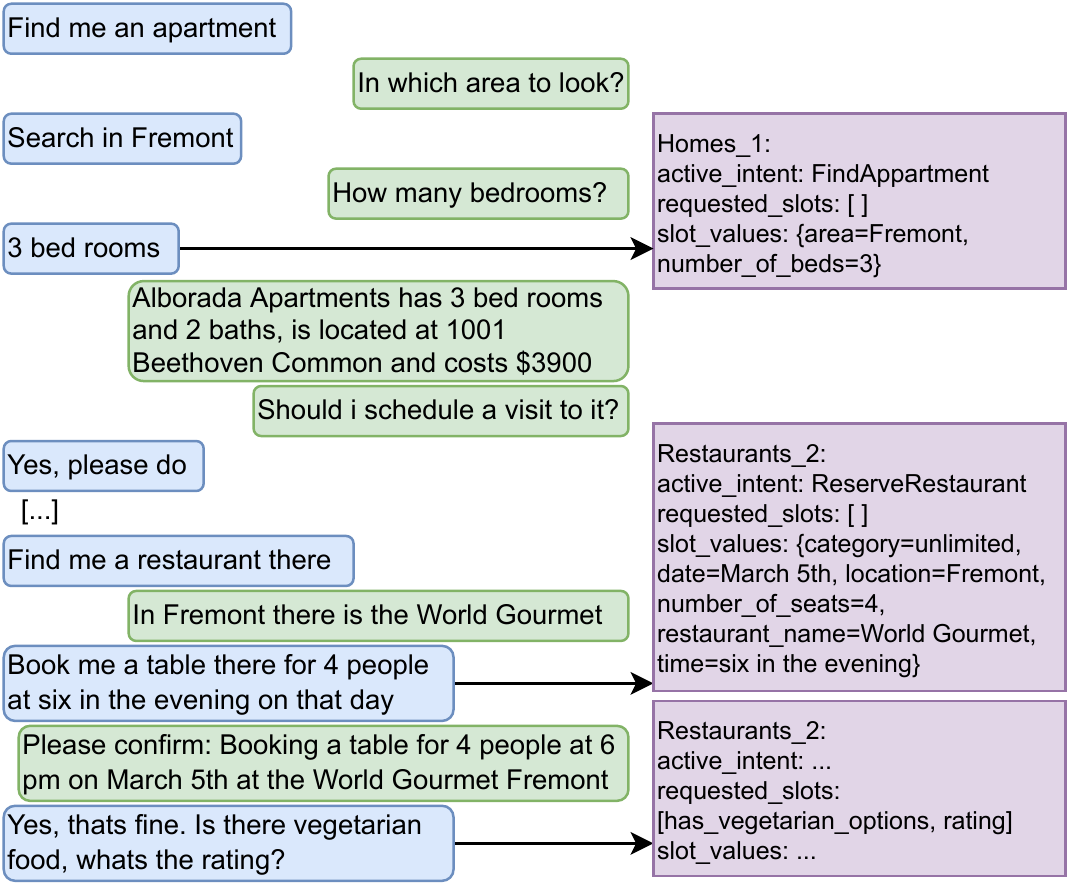}
  \caption{At every user turn the dialogue state is calculated for each involved service.
  The last two utterances are not always enough especially for the slot filling task. In such cases slot values can be found either in previous dialogue states or previous system actions. }
  \label{fig:example}
\end{figure}

\begin{figure*}[ht!]
  \centering
  \includegraphics[width=\textwidth]{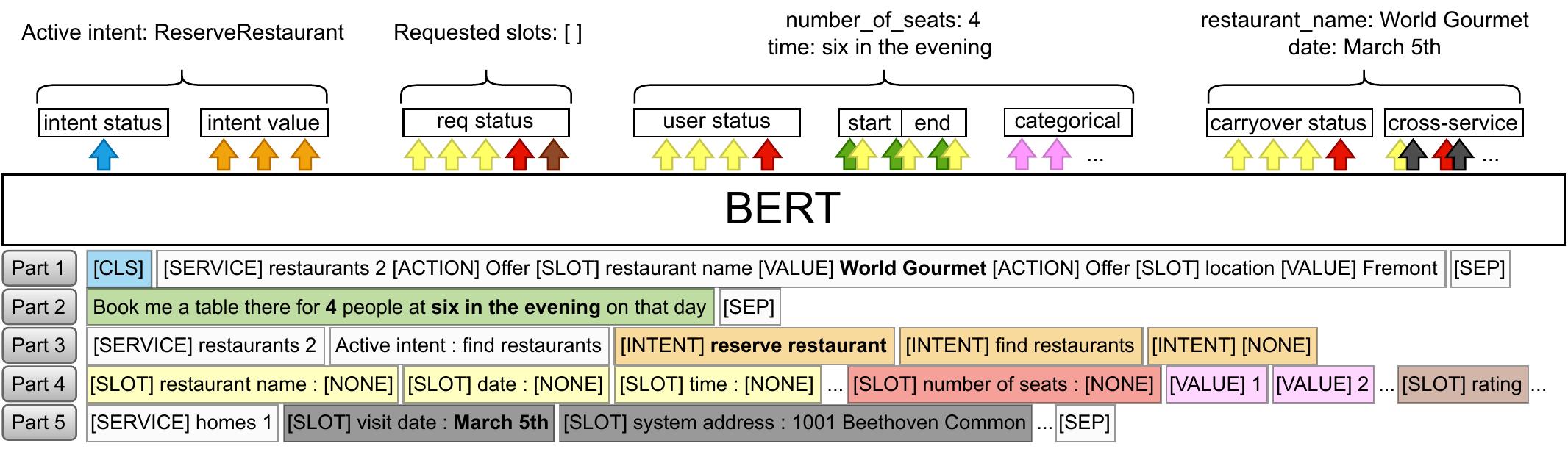}
  \caption{The inputs to the intent prediction, requested slot prediction, slot filling and slot carryover heads are shown for our proposed multi-task BERT model (top), along with an example encoding of the utterance and dialogue history that is the input to the base BERT model (bottom). Note the color coding of the input to the classification heads (top) that matches the various parts of the input sequence (bottom). For this example, 
  the service in the system and the user utterance is Restaurants\_2.
  The previous intent FindRestaurants changes to ReserveRestaurant.
  No slots are requested by the user.
  In the preceding system utterance, the system offers the value ``World Gourmet'' for the slot restaurant\_name which the user accepts (slot carryover in\_sys\_uttr).
  The user gives the values ``six in the evening'' and ``4'' for the non-categorical slot time and the categorical slot number\_of\_seats.
  The date value is not uttered but it is implied that it has been mentioned before (slot carryover in\_cross\_service\_hist from a previous service (Homes\_1)). Part of the input is truncated for illustration purposes.}
  \label{fig:model}
\end{figure*}

Pre-trained transformer models (e.g. BERT \cite{DBLP:journals/corr/abs-1810-04805}, XLNet \cite{DBLP:journals/corr/abs-1906-08237}, GPT-2 \cite{radford2019language}, T5 \cite{2020t5} etc.) are the most popular solution for schema-based DST modeling.
The importance of encoding the dialogue and schema together is highlighted in \cite{cao-zhang-2021-comparative}.
State-of-the-art methods use classification \cite{ma2020endtoend} or sequence-to-sequence pre-trained transformer models \cite{DBLP:journals/corr/abs-2109-07506, zhao-etal-2021-effective-sequence, DBLP:journals/corr/abs-2105-04222}.
The entire dialogue is passed to the model multiple times with every possible schema element description (multi-pass approach).
In \cite{zhao2022descriptiondriven}, the authors concatenate all of the schema element descriptions with the dialogue (single-pass approach) slightly improving computational efficiency, yet still using the entire dialogue history.
Other methods (\cite{ruan2020finetuning, DBLP:journals/corr/abs-2006-09035}) aim to address this issue by only encoding the last two utterances.
To retrieve slot values found in earlier utterances, slot carryover mechanisms and a multi-pass approach were employed.
Note that methods that encode the entire dialogue history, 
e.g., \cite{ma2020endtoend, zhao2022descriptiondriven}, often 
perform better than methods that only encode the last two utterances.

In this paper, we propose a single multi-task BERT-based model that jointly performs intent prediction, requested slot prediction and slot filling.
In the proposed model, we adopt slot carryover mechanisms and encode only the preceding system utterance and the current user utterance.
Furthermore, the preceding system utterance is abstracted and represented as its underlying system actions.
To achieve a more efficient and parsimonious input representation, we encode all of the schema elements together using only their names and we selectively include past dialogue states.
Our proposed model significantly outperforms the baseline SGP-DST system and achieves near state-of-the-art performance.
Extensive ablation studies reveal the impact of each strategy of our model on the slot filling task.
Our key contributions are:
1) we propose a novel multi-task BERT-based model with slot carryover mechanisms,
2) we construct an efficient and parsimonious abstracted representation of the dialogue and schema that is shown to significantly improve performance while achieving greater computational efficiency.
Our code is available as open-source \footnote{https://github.com/lefteris12/multitask-schema-guided-dst}.

\section{Method}

The multi-task model architecture is shown in Fig. \ref{fig:model}. The user utterance, previous system utterance, schema(ta) and past DST information (see Part 1 to 5) are encoded via BERT. Different pieces of the encoded sequence (see matching color coding in figure) are given as input to nine classification heads that perform the tasks of intent prediction (2 heads), requested slot prediction, slot filling (4 heads) and slot carryover (2 heads).

\subsection{Notation}
\label{sec:problem_formulation}

Let $n$ be a dialogue service, $I(n)$ the set of intents in the service (including the special \texttt{[NONE]} intent) and $S(n)$ the set of slots in the service.
Slots are divided to categorical and non-categorical slots.
Let $S_{cat}(n) \subseteq S(n)$ be the set of categorical slots and $S_{noncat}(n) \subseteq S(n)$ the set of non-categorical slots.
For every categorical slot, a set of possible values $V(s), s \in S_{cat}(n)$ are available.
Furthermore, every slot may be informable or not depending on whether the user is allowed to give a value for it.
We denote the set of the service informable slots as $S_{inf}(n) \subseteq S(n)$.

Assume that at user turn $t$ of a dialogue with $N$ services we want to predict the dialogue state for service $n$.
Essentially we have to predict the active intent $int(n)$ (intent prediction), the requested slots $req(n) \subseteq S(n)$ (requested slot prediction) and the values for the slots given by the user $usrSlotValue(s), s \in S_{inf}(n)$ (slot filling).

For every service $n', 1 \leq n' \leq N$, we denote its previous active intent as $prevInt(n')$.
Also, for every slot $s \in S(n')$ we denote the last value given by the user for $s$ as $prevUsrSlotValue(s)$.
Furthermore, we use $prevSysSlotValue(s)$ and $sysUttrSlotValue(s)$ to denote the last value present in a system action, before turn $t-1$ and at (system) turn $t-1$ respectively.
For $prevSysSlotValue(s)$ and $sysUttrSlotValue(s)$ we only use system actions that contain the slot $s$ and exactly one value for the slot.
In cases where the intent or the slot value is empty we use the \texttt{[NONE]} value.

We use $S_{prev}$ to denote the set of slots $s \in S(n'), n' \neq n$ that $prevUsrSlotValue(s)$ or $prevSysSlotValue(s)$ is not \texttt{[NONE]} and $prevSlotValue(s)$ to denote their previous value.
If $prevUsrSlotValue(s)$ is not \texttt{[NONE]} we use that value otherwise we use $prevSysSlotValue(s)$.

For every slot we employ additional binary features $x_{bin}(s)$.
The binary features used are the following: 1) whether the service is new in the dialogue 2) whether the service switched (it was not present in the previous dialogue state) 3) whether exactly one value for the slot is found in the system utterance 4) whether exactly one value for the slot is found in previous system utterances 5) whether the slot is required in at least one intent 6) whether the slot is optional in all intents.
Similar binary features have been used by \cite{ruan2020finetuning}.

\subsection{Input representation}

\label{sec:input_representation}

An example input can be seen in Fig. \ref{fig:model}.
In Part 1 we encode the preceding system utterance as a list of actions.
In Part 2 we encode the current user utterance.
In Part 3, the active service $n$, the previous active intent $prevInt(n)$ and all candidate intents belonging to service $n$ are enumerated.
Part 4 contains the list of all slots $s \in S(n)$.
If $s \in S_{inf}(n)$ we append $prevUsrSlotValue(s)$ and if $s \in S_{cat}(n) \cap S_{inf}(n)$ we also append all values in $V(s)$.
Part 5 contains all other services found earlier in the dialogue.
For every service we enumerate slot-value pairs from previous dialogue states or system actions, $s \in S_{prev}$ and their values $prevSlotValue(s)$.
We prepend the word ``system'' before slots given by the system to differentiate them from slots given by the user (present in previous dialogue states).

For the schema we only use the names for the slots and intents instead of their full natural language descriptions used by other works.
A number of custom tokens are introduced to the BERT vocabulary to indicate intents, slots etc.

\subsection{Intent prediction task}

\textbf{Intent status head}.
We perform binary classification on the encoded \texttt{[CLS]} representation to predict the intent status as active or none.

\noindent
\textbf{Intent value head}.
For every intent $i \in I(n)$ we perform binary classification on its encoded \texttt{[INTENT]} representation to predict if the user switches to that intent.

If the intent status is active we choose the intent with the highest intent value probability.
Otherwise, we keep the previous intent $prevInt(n)$.

\subsection{Requested slot prediction task}

\textbf{Requested status head}.
For every slot $s \in S(n)$ we perform binary classification on its encoded \texttt{[SLOT]} representation in Part 4 to decide whether it is requested in the current user utterance.

\subsection{Slot filling task}

\textbf{User status head}. For every slot $s \in S_{inf}(n)$ we find the user status using its encoded \texttt{[SLOT]} representation in Part 4 to decide whether a value is given in the current user utterance.
The user status classes are none, active and dontcare.

\noindent
\textbf{Categorical head}.
For the categorical slots $s \in S_{inf}(n) \cap S_{cat}(n)$ we perform binary classification for every possible value $v \in V(s)$ on its encoded \texttt{[VALUE]} representation to predict whether it is present in the user utterance.

\noindent
\textbf{Start} and \textbf{end heads}.
For the non-categorical slots $s \in S_{inf}(n) \cap S_{noncat}(n)$ we find the start and end span index distribution in the user utterance by performing classification on the concatenation of every user utterance token with the encoded \texttt{[SLOT]} representation.

If the user status is active, the value or the span with the highest probability is chosen for the slot.
If the user status is dontcare, the special dontcare value is assigned to the slot.

\subsection{Slot carryover}

The user does not always explicitly give the value for the slot but they may instead refer to previous utterances.
Therefore, we design slot carryover mechanisms to retrieve values for slots from the current or previous services.

\noindent
\textbf{Carryover status head}.
For every slot $s \in S_{inf}(n)$ we predict the carryover status using its encoded \texttt{[SLOT]} representation in Part 4 to find the source of the slot value.
The carryover status classes are none, in\_sys\_uttr, in\_service\_hist and in\_cross\_service\_hist.

For in\_sys\_uttr the slot is updated according to the value present in the preceding system utterance $sysUttrSlotValue(s)$.
For in\_service\_hist the slot is updated according to the value present in past system actions of service $n$, $prevSysSlotValue(s)$.
In the above two cases, the user accepts the value given by the system and we simply carry that value over.

\noindent
\textbf{Cross-service head}.
For every slot $s' \in S_{prev}(n)$ we perform binary classification on the concatentation of its encoded \texttt{[SLOT]} representation in Part 5 with the encoded \texttt{[SLOT]} representation of $s$ in Part 4 to decide whether we should carry the value over from slot $s'$ to slot $s$.
The highest probability slot $s'$ is used as the source for the value $s$ if the predicted carryover status is in\_cross\_service\_hist.
In this case, we assign the value $prevSlotValue(s')$ to slot $s$.

We first check the user status and if it is not none we update the slot value according to its output.
Otherwise, we also check the carryover status.
If it predicts that a carryover should take place, we update the slot value accordingly.
If both user and carryover status are none the value remains the same as in the previous dialogue state, $prevUsrSlotValue(s)$.

\subsection{Multi-task training}

For the intent status, intent value, categorical, start, end and cross-service classification heads we derive the class probabilities with a two-layer feed-forward neural network.
For the requested status, user status and carryover status classification heads we concatenate the slot binary features $x_{bin}(s)$ after the first layer.

We jointly optimize all classification heads, using the cross entropy loss for each head.
For the intent prediction task the loss is $L_1 = w_1L_{intstat} + w_2L_{intval}$, for the requested slot prediction $L_2 = L_{reqstat}$ and for the slot filling task $L_3 = w_3L_{usr} + w_4L_{carry} + w_5L_{cat} + w_6L_{start} + w_7L_{end} + w_8L_{cross}$.
Finally, the total loss is defined as $L = \lambda_1L_1 + \lambda_2L_2 + \lambda_3L_3$.

\section{Experimental Setup}

\begin{table*}[t]
  \caption{Comparison to other works}
  \label{tab:comparison}
  \centering
  \scalebox{0.8}{
  \begin{tabular}{ cccccc }
    \toprule
    \textbf{System} & \textbf{Model} & \textbf{Params}& \textbf{JGA} & \textbf{Intent Acc} & \textbf{Req Slot F1} \\
    \midrule
    SGD-baseline \cite{rastogi2020scalable} & \bertbase & \textbf{110M} & 25.4 & 90.6 & 96.5 \\
    SGP-DST \cite{ruan2020finetuning} & 6 $\times$ \bertbase & 660M & 72.2 & 91.9 & 99.0 \\
    paDST \cite{ma2020endtoend} & 3 $\times$ \robertabase $+$ \xlnetlarge & 715M & \textbf{86.5} & 94.8 & 98.5 \\
    D3ST \cite{zhao2022descriptiondriven} (Base) & \tbase & 220M & 72.9 & 97.2 & 98.9 \\
    D3ST \cite{zhao2022descriptiondriven} (Large) & \tlarge & 770M & 80.0 & 97.1 & 99.1 \\
    D3ST \cite{zhao2022descriptiondriven} (XXL) & \txxl & 11B & 86.4 & \textbf{98.8} & \textbf{99.4} \\
    Ours (median result) & \bertbase & \textbf{110M} & 82.7 & 94.6 & \textbf{99.4} \\
    Ours (avg 3 runs) & \bertbase & \textbf{110M} & 82.5 $\pm$ 1.0 & 94.7 $\pm$ 0.5 & \textbf{99.4} $\pm$ 0.1 \\
    \bottomrule
  \end{tabular}}
\end{table*}

\begin{table}[t]
  \caption{Ablation study}
  \label{tab:ablation1}
  \centering
  \scalebox{0.8}{
  \begin{tabular}{ ccccc }
    \toprule
    \textbf{System} & \textbf{JGA} & \textbf{Avg GA} \\
    \midrule
    Ours & 82.7 & 95.2 \\
    w/o system actions & 71.9 & 91.6  \\
    w. slot descriptions & 78.3 & 94.1 \\
    w/o previous state & 79.8 & 94.0  \\
    w/o schema augm. & 80.5 & 94.9 \\
    w/o schema augm. \& word dropout & 78.1 & 94.3 \\
    w/o binary features & 81.0 & 94.4 \\
    \bottomrule
  \end{tabular}}
\end{table}

\begin{table}[t]
\small
  \caption{Effect of carryover mechanisms}
  \label{tab:ablation2}
  \centering
  \scalebox{0.8}{
  \begin{tabular}{ ccccc }
    \toprule
    \textbf{System} & \textbf{JGA} & \textbf{Avg GA} \\
    \midrule
    Ours & 82.7 & 95.2 \\
    w/o in\_sys\_uttr  & 62.8 & 87.0 \\
    w/o in\_service\_hist & 76.4 & 92.7 \\
    w/o in\_cross\_service\_hist & 66.8 & 84.4 \\
    \midrule
    SGD-baseline \cite{rastogi2020scalable} & 25.4 & 56.0 \\  
    w/o in\_service\_hist \& in\_cross\_service\_hist & 61.6 & 81.9 \\
    w/o all & 36.5 & 68.5 \\
    \bottomrule
  \end{tabular}}
  
\end{table}

\noindent
\textbf{Dataset.}
We evaluate our proposed system on the SGD dataset \cite{rastogi2020scalable}.
SGD contains a total of 21,106 dialogues over 20 domains and 45 services.
We use the standard train/development/test split introduced in \cite{rastogi2020scalable}.
The test set contains 1,331 single-domain and 2,870 multi-domain dialogues and 77\% of the dialogue turns contain at least one service not present in the train set.
We use the following metrics: Joint Goal Accuracy (JGA), Average Goal Accuracy (Avg GA), Intent Accuracy and Requested Slot F1 as defined in \cite{rastogi2020scalable}.

\noindent
\textbf{Label Acquisition.}
In order to acquire labels for the user and carryover status, we use the user actions and search previous turns and dialogue states to find the source for the slot. 
We consider a slot as informable if and only if it is either required or optional in at least one intent.
For every turn we run the model only for the involved services (services with at least one change in the dialogue state in the turn) according to the ground truth dialogue states during both training and evaluation for fair comparison to other works.
The input to the model contains ground-truth previous dialogue states during training and during evaluation the previously predicted ones are used.

\noindent
\textbf{Training Setup.}
We use the huggingface \footnote{https://huggingface.co/docs/transformers/model\_doc/bert} implementation of the BERT uncased models.
For all our experiments we use a batch size of 16 and a dropout rate of 0.3 for the classification heads.
We use the AdamW optimizer \cite{DBLP:journals/corr/abs-1711-05101} with a linear warmup of 10\% of the training steps and learning rate 2e-5.
We train for a total of about 55k steps and evaluate on the development set every 4k steps.
We choose the model that performs best based on the JGA metric on the development set.

\noindent
\textbf{Preprocessing and augmentation.}
We preprocess the schema elements and the system actions by removing underscores and splitting the words when on CamelCase and snake\_case style.
We randomly ($p=0.1$) replace the input tokens in the user utterance with the \texttt{[UNK]} token (word dropout) and shuffle the order of the schema elements in Parts 3-5 during training as proposed by \cite{DBLP:journals/corr/abs-1911-03906}.
We also apply random ($p=0.1$) data augmentation through synonym replacement and random swap to the intents, slots and values in Parts 3-4 (schema augm.) via \cite{wei-zou-2019-eda}.

\section{Results and Discussion}

\noindent
\textbf{Comparison to other works.}
In Table \ref{tab:comparison} we compare our model to SGD-baseline, SGP-DST, paDST and three D3ST implementations of variable size.
The SGD baseline \cite{rastogi2020scalable} fine-tunes BERT with the last two utterances as input and uses precomputed BERT embeddings for the schema.
SGP-DST \cite{ruan2020finetuning} uses the last two utterances and slot carryover mechanisms to retrieve values for slots which were mentioned in previous utterances.
paDST \cite{ma2020endtoend} and D3ST \cite{zhao2022descriptiondriven} encode the entire dialogue history until the current turn and calculate the dialogue state from scratch.
We report the metrics and the number of parameters in the pretrained model(s) fine-tuned by each method.

Our method clearly outperforms SGP-DST in all tasks indicating that our strategies are effective.
Some of the entire-dialogue models outperform our model, especially when they use much more parameters (D3ST XXL) or apply more handcrafted features, special rules and dialogue augmentation through back-translation (paDST).
Overall, the proposed approach achieves near state-of-the-art performance despite using a much smaller model size and a shorter  input representation.

\noindent
\textbf{Ablation study.}
We perform an ablation study (Table \ref{tab:ablation1}) to show the contribution of each of the proposed strategies on the slot filling task.
Replacing the system utterance with a set of  system actions (w/o system actions) has the biggest effect on performance (see input sequence Part 1 in Fig. 2).
The system actions contain key information including the slot names and their respective values, helping our model
identify which slots are requested, offered, confirmed etc. and predict the user and carryover status most accurately.
Performance drops when we additionally include the slot descriptions for the informable slots of the current service (w. slot descriptions, see Parts 3-4 of input).
By removing previous intent and slot values in Parts 3-4 (w/o previous state) we observe a performance drop but also a training speedup because of the smaller input sequence.
We also observe an improvement by performing schema augmentation and word dropout possibly because these strategies help to avoid overfitting (w/o schema augm. \& word dropout).
The hand-crafted binary features can slightly benefit the system (w/o binary features).

\noindent
\textbf{Effect of slot carryover mechanisms.}
In Table \ref{tab:ablation2} we show the effect of the various slot carryover mechanisms.
For these experiments the model is trained once and during evaluation we replace each carryover status class with ``none''.
As expected, dropping ``in\_sys\_uttr'' has the biggest impact on performance. 
``in\_cross\_service\_hist'' is also important because of the large number of multi-domain dialogues.
By removing ``in\_service\_hist'', performance is less affected.
Without ``in\_service\_hist'' and ``in\_cross\_service\_hist'' (by only considering the last two utterances) we still achieve a higher accuracy than the SGD-baseline.

\section{Conclusions}

We propose a multi-task model for schema-guided dialogue state tracking that reasons for all three critical DST tasks simultaneously, as well as, an efficient and parsimonious encoding of user input, schemata and dialogue history. Close to state-of-the-art performance is achieved, using a significantly smaller model and input encoding. Among the various proposed enhancements to the model we show that abstracting the preceding system utterance with system actions gives
the biggest performance boost.
Strategies like appending previous dialogue states, data augmentation and adding hand-crafted features further improve performance.
We believe that these strategies can guide the design of accurate, efficient and ontology-independent task-oriented DST capable of scaling to large multi-domain dialogues, important in real world applications.

\bibliographystyle{IEEEtran}

\bibliography{mybib}

\end{document}